\documentclass[letterpaper, 10pt, journal, twoside]{IEEEtran}
\usepackage{amsmath,amsfonts,amssymb,amsthm}
\usepackage{algorithmic}
\usepackage{algorithm}
\usepackage{hhline}
\usepackage{textcomp}
\usepackage{stfloats}
\usepackage{caption}
\usepackage{url}
\usepackage{verbatim}
\usepackage{graphicx}
\usepackage{cite}

\usepackage[dvipsnames]{xcolor}
\usepackage{bm}  
\usepackage{subcaption}
\usepackage[colorlinks=true,linkcolor=blue,citecolor=blue]{hyperref} 

\hypersetup{
  colorlinks   = true,    %
  urlcolor     = blue,    %
  linkcolor    = blue,    %
  citecolor    = blue      %
}

\usepackage[T1]{fontenc}

\usepackage{booktabs}
\usepackage[normalem]{ulem}
\useunder{\uline}{\ul}{}

\newcommand{\revcolor}[1]{}

\title{Bimanual Grasp Synthesis for Dexterous \\ Robot Hands}

\begin{document}

\let\oldtwocolumn\twocolumn
\renewcommand\twocolumn[1][]{%
    \oldtwocolumn[{#1}{
        \vspace{-33pt}
        \centering
        \includegraphics[width=\linewidth]{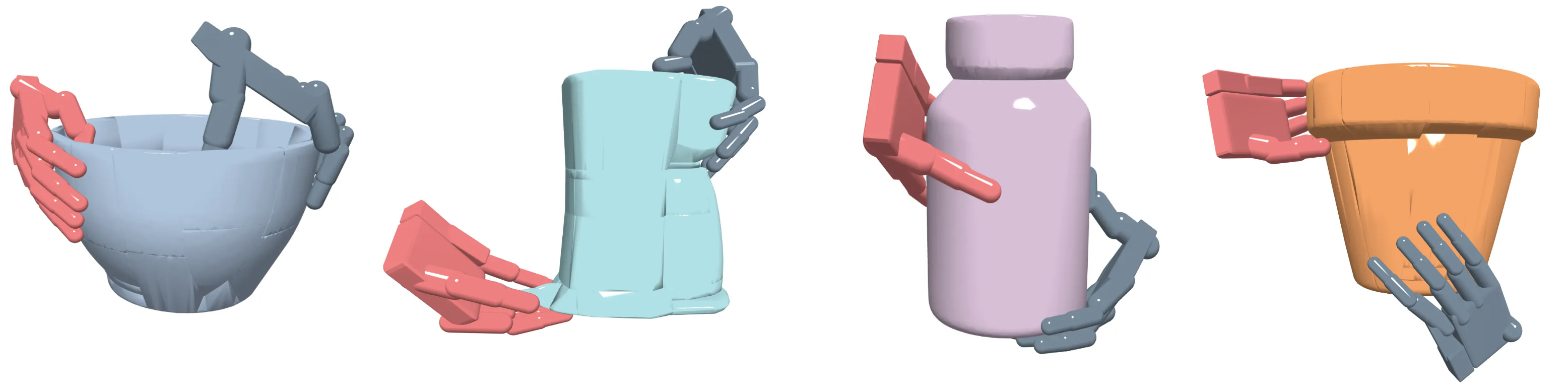}
        \captionof{figure}{Bimanual manipulation is necessary for handling large and heavy objects (e.g., basins, kitchen appliances). These objects would otherwise be difficult to handle using single-handed manipulators due to imbalanced contact forces and torques. }
        \label{fig:teaser}
        \vspace{3pt}
    }]
}

\author{Yanming Shao, Chenxi Xiao$^{*}$

\thanks{Manuscript received: Jun 29, 2024; Revised: Aug 26, 2024; Accepted: Oct 13, 2024. This paper was recommended for publication by Editor Júlia Borràs Sol upon evaluation of the Associate Editor and Reviewers’ comments.}

\thanks{This work was supported by Shanghai Frontiers Science Center of Human-centered Artificial Intelligence (ShangHAI), MoE Key Laboratory of Intelligent Perception and Human-Machine Collaboration (KLIP-HuMaCo).}
        
\thanks{Yanming Shao is with the School of Information Science and Technology at ShanghaiTech University, 
        {\tt\footnotesize shaoym2023@shanghaitech.edu.cn}}
\thanks{Chenxi Xiao ($^*$ corresponding author) is with the School of Information Science and Technology at ShanghaiTech University, 
        {\tt\footnotesize xiaochx@shanghaitech.edu.cn}}%

\thanks{Project website is available at \url{https://BimanGrasp.github.io/}.}

\thanks{Digital Object Identifier (DOI): see top of this page.}

}

\markboth{IEEE ROBOTICS AND AUTOMATION LETTERS. PREPRINT VERSION. ACCEPTED OCT, 2024 }
{Shao \MakeLowercase{\textit{et al.}}: BimanGrasp: Synthesizing Stable Bimanual Dexterous Grasps}

\maketitle

\begin{abstract}
Humans naturally perform bimanual skills to handle large and heavy objects. To enhance robots' object manipulation capabilities, generating effective bimanual grasp poses is essential. Nevertheless, bimanual grasp synthesis for dexterous hand manipulators remains underexplored. To bridge this gap, we propose the BimanGrasp algorithm for synthesizing bimanual grasps on 3D objects. The BimanGrasp algorithm generates grasp poses by optimizing an energy function that considers grasp stability and feasibility. Furthermore, the synthesized grasps are verified using the Isaac Gym physics simulation engine. These verified grasp poses form the BimanGrasp-Dataset, the first large-scale synthesized bimanual dexterous hand grasp pose dataset to our knowledge. The dataset comprises over 150k verified grasps on 900 objects, facilitating the synthesis of bimanual grasps through a data-driven approach. Last, we propose BimanGrasp-DDPM, a diffusion model trained on the BimanGrasp-Dataset. This model achieved a grasp synthesis success rate of 69.87\% and significant acceleration in computational speed compared to BimanGrasp algorithm. 
\end{abstract}


\section{Introduction}
\label{intro}

\IEEEPARstart{H}{umans} can seamlessly coordinate both hands to perform complex tasks in daily life. This is mainly due to several advantages of bimanual manipulation. For instance, bimanual manipulation enables diverse object interaction skills effortlessly, such as tying ropes, knitting clothes, and performing kitchen chores. Compared to using only a single hand, bimanual manipulation reduces human fatigue and improves body balance by distributing payloads more evenly \cite{vahrenkamp2011bimanual}. This capability is particularly beneficial when handling large and heavy objects, as it enhances both productivity and safety.

In the field of robotics, the growing market for humanoid robots has driven the development and use of multi-fingered dexterous hands for object manipulation. Recent works \cite{lundell2021ddgc,li2023gendexgrasp,wang2023dexgraspnet,li2024grasp} have mainly focused on unimanual dexterous grasping based on object geometry. These techniques have simplified the use of dexterous hands with high degrees of freedom (DoF). However, they typically focused on objects of a size and mass suitable for a single hand. This narrow focus overlooks many larger and heavier objects (e.g., heavy bottles, home appliances, and furniture) that exceed the volume of unimanual grasps. Moreover, these works involved only one dexterous hand and did not fully exploit the potential of humanoid robots, which naturally possessed two hands. 

On the other hand, there have been prior works focusing on bimanual manipulation. These works focus mainly on learning various object interaction skills. State-of-the-art works in this area have utilized Reinforcement Learning (RL) and imitation learning to acquire skills such as opening bottle lids and closing doors \cite{chen2022towards, zhang2023artigrasp, chen2023bi, fan2023arctic}. Although these studies involve bimanual grasping skills, they target very specific objects with unique functional affordances, requiring an ad-hoc training process for each object in simulation. Therefore, a research gap remains unbridged when designing general grasping skills for arbitrary objects. To our knowledge, there are no existing tools or grasp pose datasets for bimanual dexterous hand grasping. Although some studies have developed frameworks for bimanual object grasping \cite{zhai20222}, they are limited to parallel-jaw grippers. This highlights a research gap in the use of bimanual dexterous hands.

From a technical perspective, synthesizing bimanual grasp poses for dexterous hands presents greater challenges compared to the conventional grasp search problem. One of the main reasons is the significantly larger action space, which greatly increases the computational cost. For instance, the DoF of a \emph{Shadow Hand} manipulator exceed $20$ \cite{andrychowicz2020learning}. When extended to bimanual grasping, the DoF doubles, leading to a substantial increase in computational cost. In practice, reducing computational cost has been the main focus of most studies on dexterous grasp synthesis \cite{liu2021synthesizing, wang2023dexgraspnet}, where the ability to generate grasps in quasi-real time when encountering new objects is highly desirable.

In this paper, we aim to develop a bimanual grasp synthesis pipeline optimized for both grasp quality and generation speed. To achieve this, we first propose BimanGrasp algorithm, a grasp synthesis method that searches for bimanual grasp poses in a high-dimensional configuration space using stochastic optimization. Secondly, by implementing BimanGrasp algorithm with GPU-based optimization, we have synthesized the BimanGrasp-Dataset, which comprises over 150k grasps. Each grasp has been verified through simulations in the Isaac Gym environment. \cite{makoviychuk2021isaac}. The validation results demonstrate that the bimanual grasp strategy can handle large and heavy objects, which were previously unattainable with unimanual grasping techniques. Lastly, by utilizing the proposed dataset, we have significantly accelerated bimanual grasp synthesis by transforming it into a data-driven paradigm. We introduce BimanGrasp-DDPM, the first diffusion model capable of efficiently generating diverse bimanual grasp poses.

To summarize, the contributions of this paper are as follows:
\begin{itemize}
\item BimanGrasp algorithm: an offline bimanual grasp synthesizer based on stochastic optimization.
\item BimanGrasp-Dataset: A dataset of bimanual robot grasping poses, validated through physics simulation. 
\item BimanGrasp-DDPM: a quasi-real time bimanual grasp generator based on DDPM.
\item Quantitative studies comparing the performance of bimanual grasping with unimanual grasping.
\end{itemize}

\section{Related Works}
\subsection{Robot Manipulation}
\label{related:robot}
Robots are cyber-physical systems that possess the ability to interact with various objects in the physical environments. Hence, object manipulation has long been a key research area. Alone this line of research, previous works have focused on developing new skills, improving efficiency and safety. Due to the complex nature of environments, object manipulation encompasses diverse forms. One category is prehensile manipulation, where robots aim to grasp objects. The other category is non-prehensile manipulation, which primarily includes non-grasping skills \cite{billard2019trends}. Typical applications include planar pushing \cite{li2018push}, throwing and catching \cite{lan2023dexcatch}, solving a Rubik's Cube \cite{akkaya2019solving}, and playing instruments \cite{zakka2023robopianist}.

In this paper, we focus on developing bimanual object grasping skills, which belong to the prehensile manipulation category. To enable bimanual grasping, it requires obtaining cooperative grasp poses that enclose objects inside. This process is known as bimanual grasp synthesis. Previous research has extensively studied the synthesis of grasp poses, but mainly for grippers \cite{miller2004graspit}. These studies have reportedly achieved both high grasp success rates and real-time computational efficiency \cite{redmon2015real, gualtieri2016high}. However, due to the limited capability of low-DoF grippers, there is a growing need for synthesizing grasps for high-DoF manipulators, such as dexterous hands.

Synthesizing grasps for dexterous hands is more challenging than for grippers. Early researches used a grasp synthesis pipeline similar to the conventional approach for grippers. This method involves sampling grasp poses and analyzing the likelihood that each grasp pose could satisfy the force-closure condition \cite{miller2004graspit, liu2021synthesizing}. The advantage of these approaches is the ability to synthesize grasp poses for almost arbitrary object shapes. However, they are computationally expensive for manipulators with high DoF due to a significantly larger action space.

To accelerate grasp generation, recent works have adopted data-driven approaches to generate grasp poses directly \cite{xu2023unidexgrasp}. Common data-driven generative models include variational autoencoders \cite{jiang2021hand,jiang2021hand, wei2022dvgg, liu2024realdex}, normalizing flows \cite{xu2023unidexgrasp}, and diffusion models \cite{lu2023ugg, huang2023diffusion, weng2024dexdiffuser,cao2024multi}. This line of research has demonstrated improved speed and quality in the synthesis of unimanual grasps \cite{huang2023diffusion, xu2023unidexgrasp}. However, it has not yet been applied to bimanual manipulation.

\subsection{Bimanual Manipulation}

Bimanual manipulation refers to the coordinated use of both hands to manipulate objects.  This type of manipulation is essential for a wide range of activities, from daily tasks to complex professional operations. In everyday life, actions such as tying shoelaces, opening jars, and typing on a keyboard depend on the synchronized use of both hands. In professional fields, bimanual manipulation is crucial in areas like surgery and component assembly, where precision and coordination between both hands are paramount \cite{krebs2022bimanual}.

Bimanual manipulation is also a critical skill for robots. Historically, humanoid robots capable of bimanual object manipulation emerged in the 2000s \cite{platt2006learning, smith2012dual}. Since then, bimanual skills such as moving kitchen cookware \cite{vahrenkamp2010integrated}, dishwashing \cite{bohg2012task}, and object pick-and-place \cite{saut2010planning} have been developed. However, these early strategies were based on two parallel grippers rather than dexterous hands.

Recent advancements in robotics have increasingly focused on learning dexterous bimanual manipulation with multi-fingered hands, often using human demonstrations. This progress has enabled robots to perform coordinated actions with greater precision than unimanual approaches \cite{fan2023arctic, zhang2023artigrasp, lin2024learning}. Despite these advancements, current researches on dexterous bimanual manipulation target very specific objects and lack large-scale datasets or diverse hand pose types \cite{feix2015grasp, harada2012pick}. To bridge the gap, we propose techniques and a dataset that aim to help humanoid robots develop more general bimanual grasping skills.

\section{Method}

\label{sec:method}

\subsection{Problem Definition and Overview} \label{method:problem_def}

\begin{figure*}[t]
    \centering
    \includegraphics[width=\linewidth]{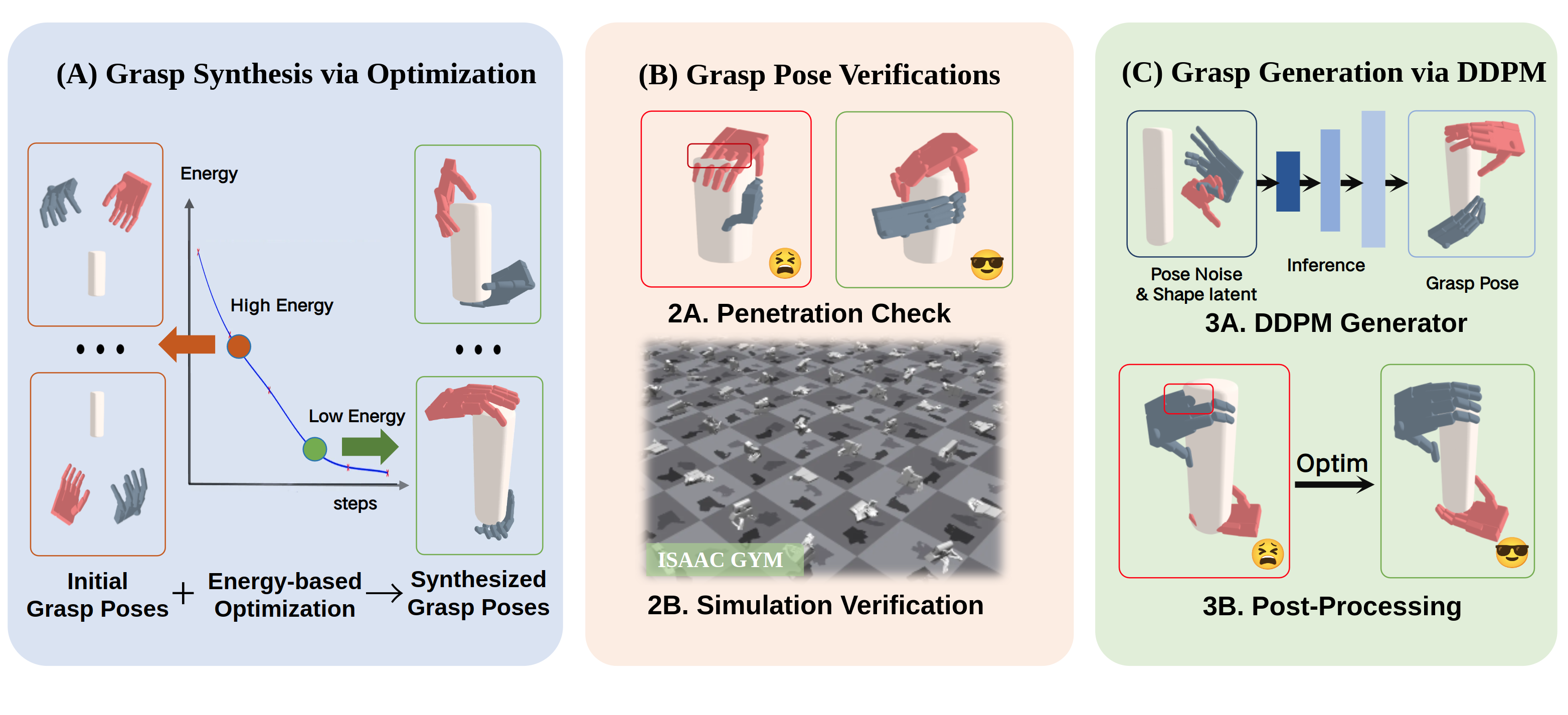}
    \caption{Our pipeline for synthesizing stable bimanual grasps, which includes: (A) generating grasp poses by initializing the bimanual grasp poses around the objects and then improving their quality through optimization; (B) verifying the grasp poses based on shape penetration and physics simulation using Isaac Gym; (C) utilizing the verified grasps (BimanGrasp-Dataset) to train a generative model (BimanGrasp-DDPM), with post-processing techniques to remove penetrations. }
    \label{fig:architecture}
\end{figure*}

{\revcolor{red}
The problem of bimanual grasp synthesis is formulated as Eq.~(\ref{eq:problem_defition}). Given an object mesh denoted as $O$, our goal is to obtain grasp poses by maximizing the grasp quality score $S$ that is empirically defined. The metric $\mathcal{G}$ used for calculating the score also considers rigid body poses $\bm{T}_l \in SE(3)$ and joint configurations $\bm{\theta}_l \in \mathbb{R}^{22}$, where $l \in \{1, 2\}$ denotes the left and right manipulators, respectively. This optimization paradigm is employed for synthesizing bimanual grasps, corresponding to the BimanGrasp algorithm proposed in Sec.~\ref{method:sec_b}.
}

\begin{equation}
\max_{\bm{T}_1, \bm{\theta}_1, \bm{T}_2, \bm{\theta}_2} S = \mathcal{G}(O, \bm{T}_1, \bm{\theta}_1, \bm{T}_2, \bm{\theta}_2).
\label{eq:problem_defition}
\end{equation}

Although this approach has been commonly adopted for grasp synthesis problems involving manipulators with lower DoF (e.g., \cite{dai2018synthesis, liu2021synthesizing}, to mention a few), one issue lies in computational efficiency. Given the high DoF of two dexterous hands, obtaining feasible bimanual grasps in quasi-real time is difficult. This limitation hinders the applications of such methods, especially for scenarios where timely responses are crucial.  

Compared to the optimization-based paradigm discussed above, synthesizing bimanual grasps through deep learning models could be more efficient. To verify this idea, we propose BimanGrasp-DDPM model (Sec.~\ref{method:sec_d}). Nevertheless, training such a generative model requires a dataset of bimanual grasp poses, which does not currently exist. To bridge this gap, we introduce the BimanGrasp-Dataset, which is synthesized offline using the optimization-based BimanGrasp algorithm. The overall system architecture for achieving this is shown in Fig.~\ref{fig:architecture}. 

\subsection{Synthesis Bimanual Grasp via Stocastic Optimization}
\label{method:sec_b}

This section describes the BimanGrasp algorithm, which is capable of synthesizing grasp poses conditioned on object meshes. The algorithm consists of two steps: 1) initialize grasp poses around the target object, and 2) iteratively optimize an energy function, during which the poses are adjusted based on grasp stability and penetration. The detailed procedures for achieving this are as follows:

{\revcolor{red}

\textbf{Step 1: Initialize Bimanual Hand Poses}. Humans naturally grasp objects by facing them and approaching from two opposite sides. In our algorithm, we empirically initialize each hand's pose symmetrically around the object's center to increase the similarity to human grasps and reduce the chance of penetration. Specifically, we adopt the initialization procedures from \cite{wang2023dexgraspnet}. First, two hands are placed on an inflated convex hull enveloping the object, with the palms facing the object. Then, we progressively decrease the hull's size, reducing the gap between the hands and the object until they make contact. Note that it's important to introduce variations in the initial hand poses and joint angles. This randomization technique enlarges the search space, allowing for more diverse grasps and helping avoid sub-optimal local minima.

}

\begin{figure}[htbp]
    \centering
    \includegraphics[width=\linewidth]{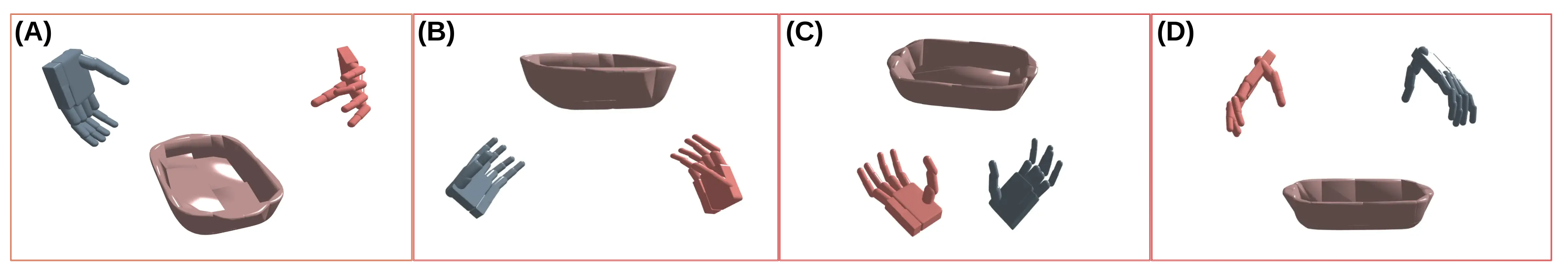}
    \caption{{\revcolor{red} Initialization of grasp poses (with randomization applied to joint angles and poses). We demonstrate four examples, which approach the object from different directions. }}
    \label{fig:initialization}
\end{figure}

\begin{table}[]
\begin{center}
\caption{Energy function for grasp search problem. The minimization objective of the algorithm is the weighted sum of all terms.}
\label{tbl:energy}
\begin{tiny}
\begin{tabular}{cc}
\hline\hline \\[-4pt]

{\footnotesize \textbf{Term}} & {\footnotesize \textbf{Formulation} }   \\[2pt]       \hline \\ 
$E_{\text{dis}}$: Hand-object distance                       &  $ \sum \limits_{a=1}^{n} d(x_a,O)$ \\[8pt]

$E_{\text{fc}}$: Force Closure                              & $||\boldsymbol{G} \boldsymbol{c}||_2$  \\[8pt]

$E_{\text{vew}}$: Wrench Ellipse Volume                     &   $  \left( \det\left(\mathbf{G} \mathbf{G}^T\right) \right)^{-\frac{1}{2}} $    \\[12pt]

$E_{\text{objpen}}$: Hand-Object Penetration                &    $\sum \limits_{l \in \{1,2\}} \sum \limits_{p_l \in   P(H_l)} \max(\delta-d(p_l,O),0)$                   \\[12pt]

$E_{\text{selfpen}}$: Hand Self-Penetration                 &    $\sum \limits_{l \in \{1,2\}} \sum \limits_{p,q \in P(H_l)} \max (\delta-d(p, q), 0)$ \\ [12pt]

$E_{\text{bimpen}}$: Inter-Hands Penetration         &    $ \sum \limits_{p \in P(H_1), q \in P(H_2)} \max (\delta-d(p, q), 0)$   \\ [12pt]

$E_{\text{joint}}$: Violation of Joint Limits                 &   $  \sum \limits_{i=1}^{44} ( \max (\theta_i - \theta_i^{max}, 0) + \max ( \theta^{min} - \theta_i, 0 ) ) $    \\ \\
\hline\hline 

\end{tabular}

\end{tiny}
\end{center}
\end{table}

\textbf{Step 2: Improve Grasp Quality}. The grasp quality is improved by optimizing an energy function, defined as the weighted sum of all terms in Table.~\ref{tbl:energy}. In the table, we denote $d(p,q)$ for $p,q \in \mathbb{R}^3$ as the Euclidean distance between two points, and  $d(p,O) = \min \limits_{q \in O}(p,q)$ as the distance between $p$ and the object mesh. We denote $H_l$ as the mesh of each hand with $l \in \{1,2\}$, and $P(H_l)$ as the anchor points selected from the hand mesh to compute penetration.

{\revcolor{red}
The empirical quality metric considers both the grasp stability and feasibility. One main goal is to keep hands close to object's surface. To achieve this, we construct the term $E_{\text{dis}}$, which quantifies the distance between the two hands and the object. By minimizing this term, the fingers and palms land close to the object surface. Here $x_a$ with $a \in \{1,2,\dots,n\}$ denotes a point cloud with $n=4000$ points sampled from both hands' surfaces.

The term $E_{\text{fc}}$ represents the force closure, which serves as the main heuristic for grasp stability. In $E_{\text{fc}}$, $c$ is the contact normal vector at the contact points $\bm{x}_j = (x_j, y_j, z_j)$, where $j \in \{1,2,\dots,8\}$ is the index of contact points. Note that unlike unimanual grasping, our approach leverages $8$ contact points for grasp collaboration ($4$ from each hand). This formulates the grasp matrix $\bm{G}$:}

\begin{equation}
\begin{aligned}
\boldsymbol{G}= & {\left[\begin{array}{ccc|ccc}
\boldsymbol{I} & \dots & \boldsymbol{I} & \boldsymbol{I}   & \dots & \boldsymbol{I}      \\
\boldsymbol{R}_{1} & \dots & \boldsymbol{R}_4 & \boldsymbol{R}_5 & \dots & \boldsymbol{R}_8
\end{array}\right] } 
\end{aligned}
\label{eq:force_closure}
\end{equation}

where

\begin{equation}
\begin{aligned}
{\revcolor{red}\boldsymbol{R}_j=\left[\begin{array}{ccc}
0 & -z_j & y_j \\
z_j & 0 & -x_j \\
-y_j & x_j & 0
\end{array}\right]}, \mathbf{I}=\left[\begin{array}{lll}
1 & 0 & 0 \\
0 & 1 & 0 \\
0 & 0 & 1
\end{array}\right].
\end{aligned}
\end{equation}

In addition,  $E_{\text{vew}}$  is defined to prevent the Gram matrix $\bm{G}\bm{G}^T$ from being ill-conditioned. This ensures that the grasp can effectively resist small external wrench disturbances from any direction. Notably, optimizing $E_{\text{fc}}$ and $E_{\text{vew}}$ establishes the differential force closure condition proposed by \cite{liu2021synthesizing}.

We also prevent penetration failures by accounting for the following penetration patterns: (1) between two hands and the object; (2) between the left and right hands; and (3) within each hand and itself. The energy terms that prevent these three types of penetrations are $E_{\text{objpen}}$, $E_{\text{bimpen}}$, and $E_{\text{selfpen}}$, respectively. Each of these energy terms is calculated with the distance between some anchor points from $P(H_1)$ and $P(H_2)$, unless it is lower than a fixed small threshold $\epsilon$.

In addition, an energy term $E_{\text{joint}}$ is introduced to handle joint limit violations. For each joint angle $\theta_i$ with $i \in \{1,2,\dots,44\}$, we denote $\theta^{\text{min}}_i$ as its lower limit and $\theta^{\text{max}}_i$ as its upper limit. If it is outside $ [ \theta_i^{\text{min}}, \theta_i^{\text{max}} ]$, the violation is penalized using the out-of-range value, in form of $\max (\theta_i - \theta_i^{\text{max}}, 0) + \max ( \theta^{\text{min}} - \theta_i, 0 ) $.

\begin{figure}[htbp]
    \centering
    \includegraphics[width=.99\linewidth]{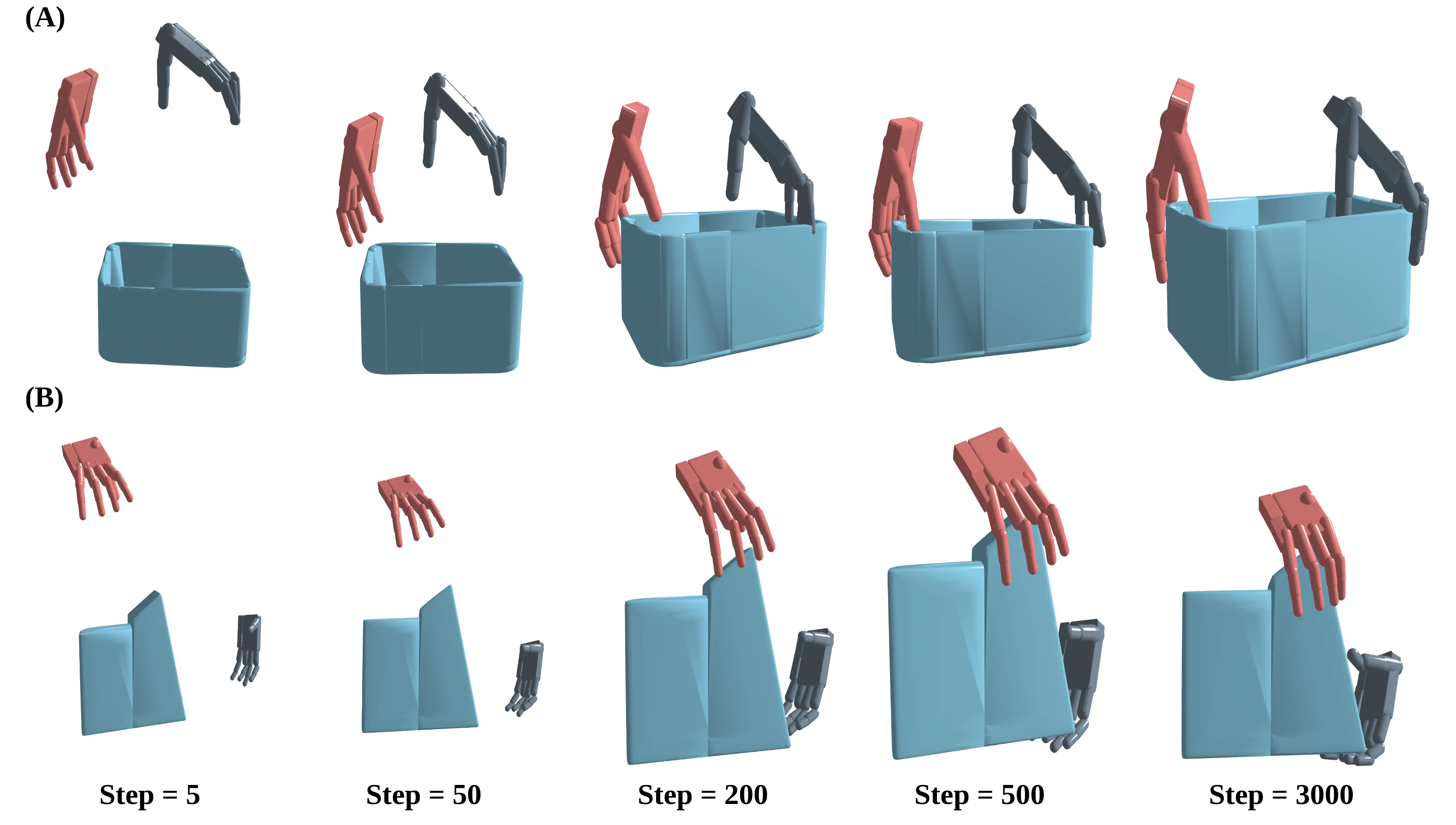}
    \caption{Visualization of the BimanGrasp algorithm's optimization process on two objects: (A) a square container, and (B) a home appliance. Both hands started by approaching the object from initialized poses. As optimization proceeds, the hands gradually landed on object surfaces.}
    \label{fig:optimization}
\end{figure}

We jointly optimize the weighted sum of all the aforementioned energy terms. Given the non-convexity of the energy function, we employ the Metropolis-adjusted Langevin algorithm (MALA) optimizer, which introduces stochasticity to circumvent local optima \cite{liu2021synthesizing}. The hand configurations during optimization process are showcased in Fig.~\ref{fig:optimization}. 

\subsection{Dataset Generation}
\label{method:sec_c}

To prepare data for training generative models, we synthesized grasp poses for a dataset of objects from Google's Scanned Objects (GSO) Dataset \cite{downs2022google}. The manipulator used is a pair of \emph{Shadow Hand}s. Each \emph{Shadow Hand} has $22$ actuated joints (denoted as $\bm{\theta}_i$ each), and a $6$ dimensional rigid body pose. For both manipulators, our action space has $(22+6) \times 2 = 56$ dimensions in total. 

{\revcolor{red}
We followed the grasp synthesis procedures outlined in Sec.~\ref{method:sec_b} to generate an initial set of grasp poses. Then, we used the Isaac Gym environment \cite{makoviychuk2021isaac} to label whether a grasp can successfully lift and hold objects, as shown in Fig.~\ref{fig:gym_valid}. The friction coefficient for both the objects and hands are fixed at $3$ (same as \cite{wang2023dexgraspnet,lu2023ugg}). To control the motor output, we employed a PD controller in Isaac Gym, with stiffness $K_p=1000.0$ and damping $K_d=10.0$.

 \begin{figure}[htbp]
     \centering
     \includegraphics[width=.95\linewidth]{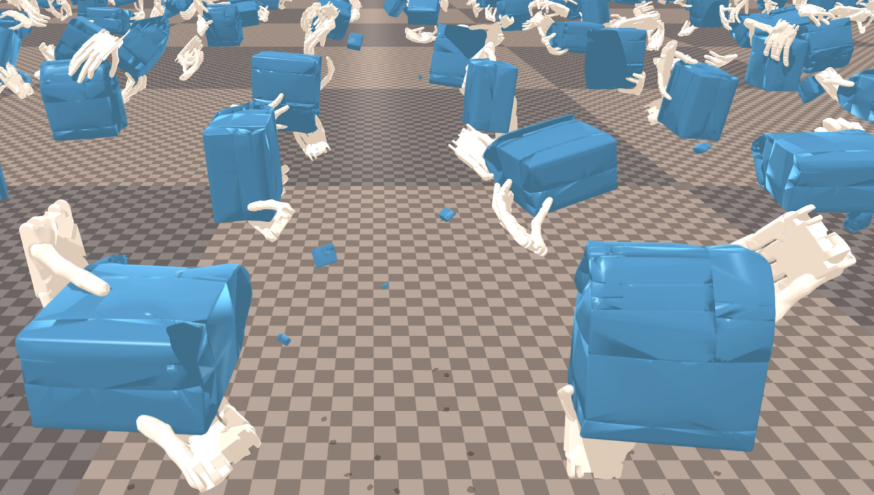}
          \caption{The visualization of our physical verification of the bimanual grasps using Isaac Gym \cite{makoviychuk2021isaac}. Objects are held firmly by stable grasps, while they slip away from unstable grasps.}
     \label{fig:gym_valid}
 \end{figure}
 
A grasp configuration is labeled as successful if the object remained in the hand for $2.0$ seconds of simulation time (\emph{i.e.}, $120$ steps at $60 \text{ Hz}$) over $6$ evaluation trials under a gravity of $9.8 \text{ m}\cdot\text{ s}^{-2}$. During each evaluation, the object and hands were randomly rotated together to verify the grasp under varying gravity force directions.}

We also checked for the three types of penetration described in Sec.~\ref{method:sec_b}. If the total penetrations exceeded $1.5 \text{ mm}$, the grasp configuration fails the evaluation. All successful grasps, along with the objects they were conditioned on, were saved into our BimanGrasp-Dataset.

\subsection{Grasp Generation Through Data Driven approach}
\label{method:sec_d}

{\revcolor{red}
We then developed a generative model based on the dataset of successfully grasped poses. The proposed generative model employed the Denoising Diffusion Probabilistic Models (DDPM) architecture \cite{ho2020denoising}, which is widely used for various generative tasks. Conditioned on a feature vector ${\revcolor{red}\mathcal{O}} \in \mathbb{R}^{1024}$ (extracted by PointNet \cite{qi2017pointnet} from a point cloud sampled from $O$), the DDPM model aims to transform standard Gaussian noise $\bm{h}^T \sim \mathcal{N}(0;I) $ into a grasp pose $\bm{h}^{0}$. This was achieved through an iterative denoising process, as described in Eq.~(\ref{eq:DDPM}):

\begin{align}
    p_{\theta} (\bm{h}^{0}|{\revcolor{red}\mathcal{O}}) &= p(\bm{h}^T) \prod \limits_{t=1}^T p(\bm{h}^{t-1}|\bm{h}^t,{\revcolor{red}\mathcal{O}}) 
    \label{eq:DDPM}
\end{align}
}

For each stage of the denoising process, the implementation was based on the reparameterization trick \cite{kingma2013auto}, as given in Eq.~(\ref{eq:DDPM_denoise}). We leveraged the U-Net model to predict the mean $ \mu_{\theta} $ and the standard deviation $ \bm{\Sigma}_{\theta}$ of the noise added at each diffusion step.

\begin{align}
    p(\bm{h}^{t-1}|\bm{h}^t,{\revcolor{red}{\revcolor{red}\mathcal{O}}}) &= \mathcal{N} (\bm{h}^{t}; \bm{\mu}_{\theta} (\bm{h}^t, t, {\revcolor{red}{\revcolor{red}\mathcal{O}}}), \bm{\Sigma}_{\theta} (\bm{h}^t, t, {\revcolor{red}{\revcolor{red}\mathcal{O}}}))
    \label{eq:DDPM_denoise}
\end{align}

Then, the learning objective is given in Eq.~(\ref{eq:DDPM_loss}):

\begin{equation}
    L_{\theta} (\bm{h}^{0}| {\revcolor{red}\mathcal{O}}) = E_{t, \epsilon, \bm{h}^{0}} \left[ || \epsilon - \epsilon_{\theta} ( \sqrt{\overline{a_t}} \bm{h}^{0} + \sqrt{1-\overline{a_t}}\epsilon, t, {\revcolor{red}\mathcal{O}} ) || \right].
    \label{eq:DDPM_loss}
\end{equation}

where $ \epsilon $ is the noise to estimate; $\epsilon_{\theta}$ is the noise predicted by the model; $ t $ represents the denoising steps; $\overline{a_t} $, defined as $ \prod \limits_{k=1}^t a_k $, represents the noise intensity.

Nevertheless, grasp poses directly obtained from the DDPM model could be infeasible. This is mainly because penetrations are not explicitly considered during the generative process. While preventing DDPM from generating penetrated grasps requires nontrivial customization, we leverage a post-processing technique. That is, after obtaining the grasp poses from DDPM, we improve the poses by optimizing on energy terms defined in Table \ref{tbl:energy}. We use much fewer ($100$) steps, compared to $10000$ steps in BimanGrasp Algorithm, to keep the computational efficiency.

\section{Experiments}
\label{section:experiments}

To validate our proposed approaches, we conducted experiments to: 1) evaluate the performance of the BimanGrasp algorithm quanlitatively and quantitatively (Sec.~\ref{section:eval_bim}); 2) evaluate the performance of the DDPM trained on the BimanGrasp-Dataset (Sec.~\ref{exp:ddpm}). Last but not least, 3) we provide discussions on experiments, ablation studies, and broader insights (Sec.~\ref{exp:discussions}).

\subsection{Evaluation on Bimanual Grasp Synthesis} 
\label{section:eval_bim}
\begin{figure}
        \centering
        \includegraphics[width=\linewidth]{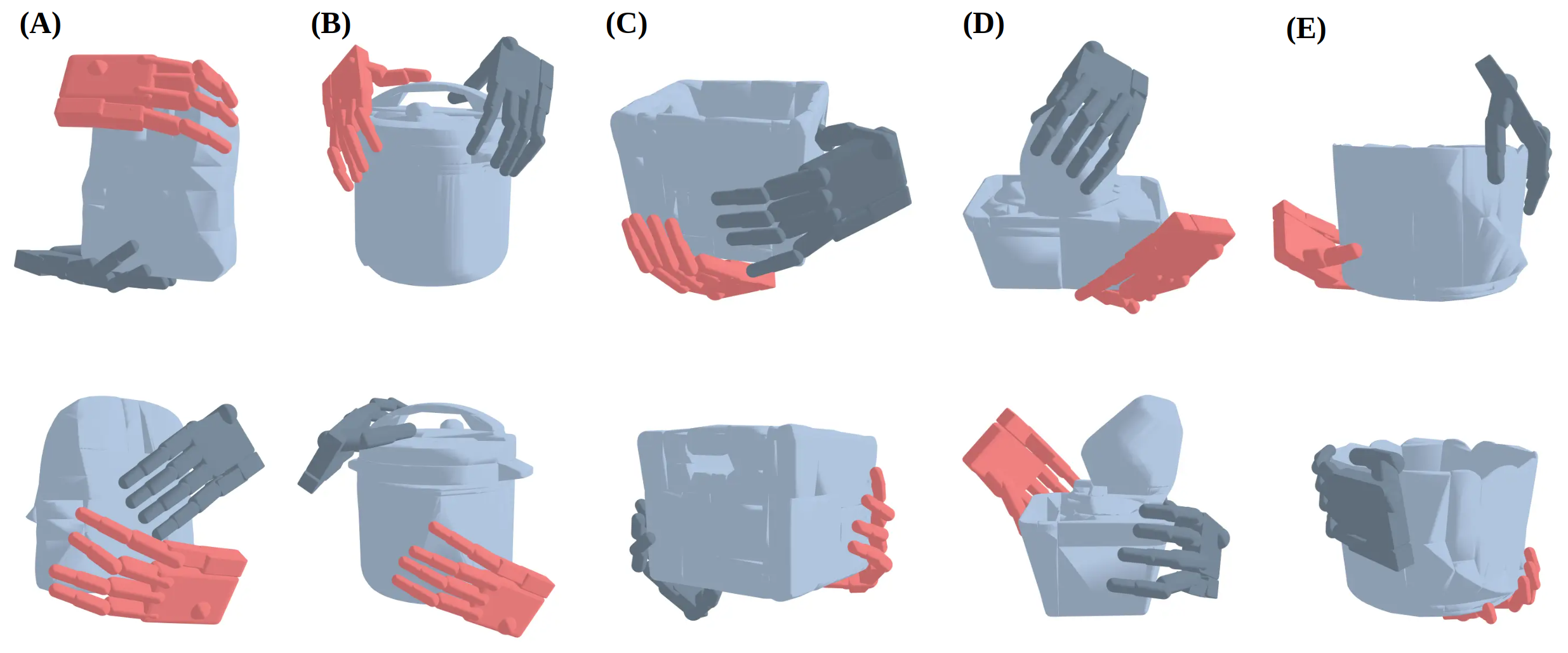}
        \caption{Visualization of the grasp poses synthesized on daily-life objects using BimanGrasp algorithm. Objects include: (A) backpack, (B) pot, (C) box, (D) bucket, and (E) container. All object models are from Google Scanned Dataset \cite{downs2022google}.}
        \label{fig:exp_household}
\end{figure}

{\textbf{Visualization of Grasps}.} 
First, we evaluate our proposed BimanGrasp algorithm on everyday objects from the GSO \cite{downs2022google} dataset. These objects include various household items, such as containers and kitchen utensils, as shown in Fig.~\ref{fig:exp_household}. The generated grasps for these daily-life objects are diverse. For instance, some grasps secure the body of a cylindrical container, while others pinch the edges of a box. These results successfully demonstrate the reliable generation of grasp poses for objects with diverse shapes.

{\revcolor{red} One question is whether the generated grasps are human-like \cite{wang2023dexgraspnet,liu2024realdex}. To validate this, we adopted the evaluation approach used in \cite{liu2024realdex}. We employed \emph{GPT-4 Vision} to score each bimanual grasp on a scale of 1 to 3 points (evaluating 1,000 grasps, with 3 views per grasp). The average score obtained was 2.67.}

{\textbf{Grasp Success Rate}.} Next, we quantitatively evaluate the quality of the grasp poses synthesized by BimanGrasp. The evaluation was conducted on the synthesized $450$k bimanual grasps ($900$ objects, with $500$ poses per object). A grasp pose is considered successful if it meets two criteria: 1) no penetration occurs, and 2) object does not slip away during the physics verification process, as outlined in Sec.~\ref{method:sec_c}. During physics verification, all $900$ objects were used in their original sizes provided in \cite{downs2022google}, with a density of $\rho = 2500 \text{ kg} \cdot \text{m}^{-3}$ {\revcolor{red} and object friction coefficient of $3$ (following setting \cite{wang2023dexgraspnet,lu2023ugg}). }

\begin{figure}[htbp]
    \centering
    \includegraphics[width=.99\linewidth]{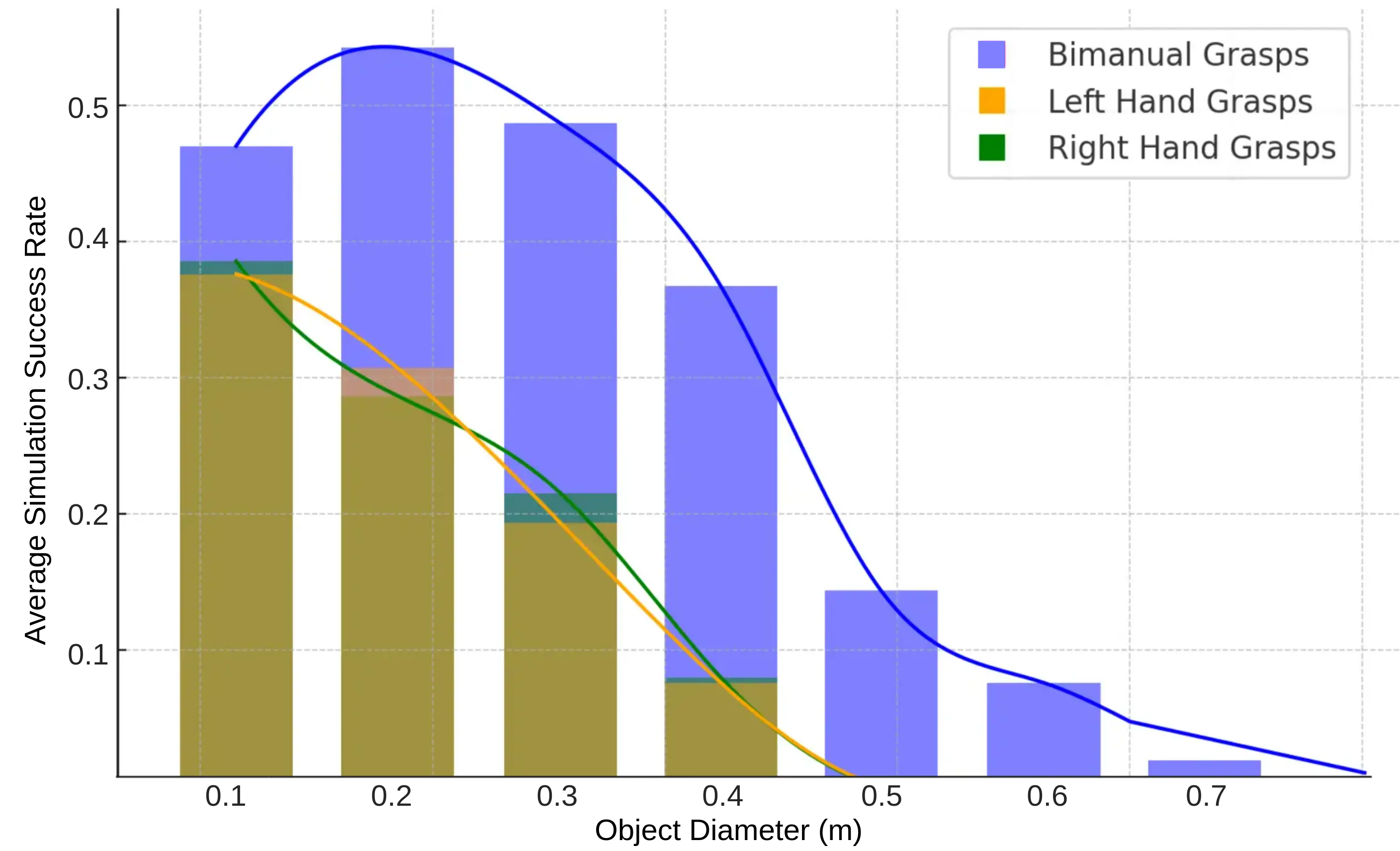}
    \caption{The average grasp success rate across various object diameters. As object diameters increase, the  success rates decrease due to the increased grasp difficulty. In all experiments, bimanual grasps outperform unimanual baselines.}
    \label{fig:radius_succ_rate}
\end{figure}

Our experimental results visualize the relationship between the object's diameter ($d$) versus grasp success rate, as illustrated in Fig.~\ref{fig:radius_succ_rate}. We compared our bimanual grasping strategy with two unimanual grasp baselines from a current state-of-the-art approach \cite{wang2023dexgraspnet}. After benchmarking all grasp poses, we visualized the success rate distribution in Fig.~\ref{fig:radius_succ_rate}. Objects were grouped into seven categories based on their diameters. The results show that our bimanual grasping strategy consistently achieves a higher success rate than unimanual strategies across all object sizes. Furthermore, the performance advantage of the bimanual strategy increases with object diameter. Unimanual grasps almost entirely fail for objects larger than $d=0.5$ m, while bimanual grasping remains effective for objects up to $d=0.7$ m. This highlights the superior effectiveness of bimanual grasps, especially for larger objects.

\begin{table}[h]
\centering
\caption{Comparison of the average grasp success rate  ($\%$) under different object densities ($\textit{kg} \cdot \textit{m}^{-3}$). }
\label{table:density}
\resizebox{.9\linewidth}{!}{
\begin{tabular}{@{}cccc@{}}
\toprule
\textbf{Density} & $\rho = 5000$ & $\rho = 2500$ & $\rho = 500$ \\ \midrule
Both hands          & $\mathbf{41.02}$ & $\mathbf{54.03}$ & $\mathbf{71.42}$ \\ 
{\revcolor{red} Uni2Bim (opt)}      & {\revcolor{red} $32.87$}  & {\revcolor{red} $45.26$} & {\revcolor{red} $56.69$} \\
Left Hand Only      & $23.38$ & $41.48$ & $68.42$ \\ 
Right Hand Only     & $21.85$ & $41.95$ & $68.48$ \\ 
\bottomrule
\end{tabular}
}
\end{table}

Next, we evaluated the grasp success rates for objects of different masses. To minimize the influence of object size, all objects were normalized to diameter $d=0.2 \text{ m}$. Then, we created three comparative groups by varying the object density: $\rho = 5000, 2500, \text{and } 500 \text{ kg} \cdot \text{m}^{-3}$. Within each group, we evaluated the average grasp success rate across all $900$ objects in the Isaac Gym simulation. 

From results in Table.~\ref{table:density}, we conclude that bimanual grasping offers significant advantages in object manipulation, particularly when handling heavy objects. Across all groups, bimanual grasp strategies consistently achieved higher success rates than unimanual strategies. This advantage is especially significant when grasping heavy objects. For instance, when $\rho = 5000 \text{ kg} \cdot \text{m}^{-3}$, the bimanual grasp strategy achieved a success rate of $41.02\%$, while the unimanual grasp strategies only achieved $23.38\%$ and $21.85\%$. {\revcolor{red} Note that this advantage is not solely due to the larger forces generated by more motors from two hands, but also because of the cooperation between them. To demonstrate this, we implemented a baseline named Uni2Bim (opt) for comparison with the BimanGrasp algorithm. Uni2Bim (opt) optimizes each hand separately, following \cite{wang2023dexgraspnet}, without any joint optimization between the two hands. As shown in results, Uni2Bim (opt) achieved lower success rates across all three object densities, highlighting the importance of joint optimization architecture. }

{\revcolor{red}
In addition, we evaluate the grasp robustness by varying the friction coefficient. To do this, we fix the object density at $2500 \ \textit{kg} \cdot \textit{m}^{-3}$ and allow the object friction coefficient to range from $0.5$ to $3.0$ (following \cite{liu2024visual}). The overall simulation success rate is shown in Table~\ref{table:friction}. When the friction coefficient was significantly reduced to 0.5, the grasp success rate decreased to 45.40\% from 54.03\%, indicating that around 84\% of all grasps are still valid. This demonstrates the robustness of our generated grasps.

\begin{table}[h]
\centering
\caption{{\revcolor{red} Success rate ($\%$) of BimanGrasp in IsaacGym under different friction coefficient settings.  Object density is at $\rho = 2500 \text{ kg} \cdot \text{m}^{-3}$. }}
\label{table:friction}
\resizebox{.99\linewidth}{!}{
\begin{tabular}{@{}ccccccc@{}}
\toprule
\textbf{Friction Coeff.} & $0.5$ & $1.0$ & $1.5$ & $2.0$ & $2.5$ & $3.0$ \\ \midrule
\textbf{Succ. Rate} & 45.40 & 47.04 & 49.32 & 51.14 & 52.44 & 54.03 \\
\bottomrule
\end{tabular}
}
\end{table}
}

\subsection{Evaluation on the Bimanual DDPM} \label{exp:ddpm}

 Using the synthesized dataset, we trained a DDPM model following procedures described in Sec.~\ref{method:sec_d}. The training was performed on the successful grasps from BimanGrasp-Dataset, (randomly selected $900 \times 75\% = 675$ objects). We then evaluated the grasping success rate on the remaining $25\%$ of objects ($225$ unseen objects) using Isaac Gym with $500$ grasps per object. The performance of the synthesized grasps for unseen objects was evaluated under a uniform diameter of $d=0.2 \text{ m}$ and varying object densities $\rho = 5000, 2500$, and $500 \text{ kg} \cdot \text{m}^{-3}$. The average success rate was $42.39\%$ for $\rho = 5000 \text{ kg} \cdot \text{m}^{-3}$, $54.06\%$ for $\rho = 2500 \text{ kg} \cdot \text{m}^{-3}$, and $69.87\%$ for $\rho = 2500 \text{ kg} \cdot \text{m}^{-3}$. These results are comparable to the success rates of the analytically synthesized bimanual grasp poses from BimanGrasp algorithm, as shown in Table.~\ref{table:density}. 
 
{\revcolor{red}
Next, we aimed to assess whether the performance of Bimanual-DDPM is comparable to other methods. Since no existing methods currently address bimanual grasp synthesis, we manually crafted two baselines. (1) We customized a Conditional Variational Autoencoder (CVAE) \cite{jiang2021hand} and re-trained the network using a bimanual grasping protocol based on the BimanGrasp dataset. (2) We used the approach from \cite{lu2023ugg} to generate grasps for each hand separately (referred to as Uni2Bim (dm)), without utilizing our dataset priors that account for hand collaboration. When evaluated on objects with a density of $2500 \text{ kg} \cdot \text{m}^{-3}$, CVAE achieved a success rate of 11.85\%, while Uni2Bim (dm) reached 36.52\%. Both baselines performed significantly worse than BimanGrasp-DDPM, although CVAE has advantages in its running time (only 18 ms per grasp).}

\begin{figure}[htbp]
    \centering
    \begin{subfigure}{0.49\textwidth}
        \label{fig:grasps_sub1}
        \centering
        \includegraphics[width=0.97\linewidth]{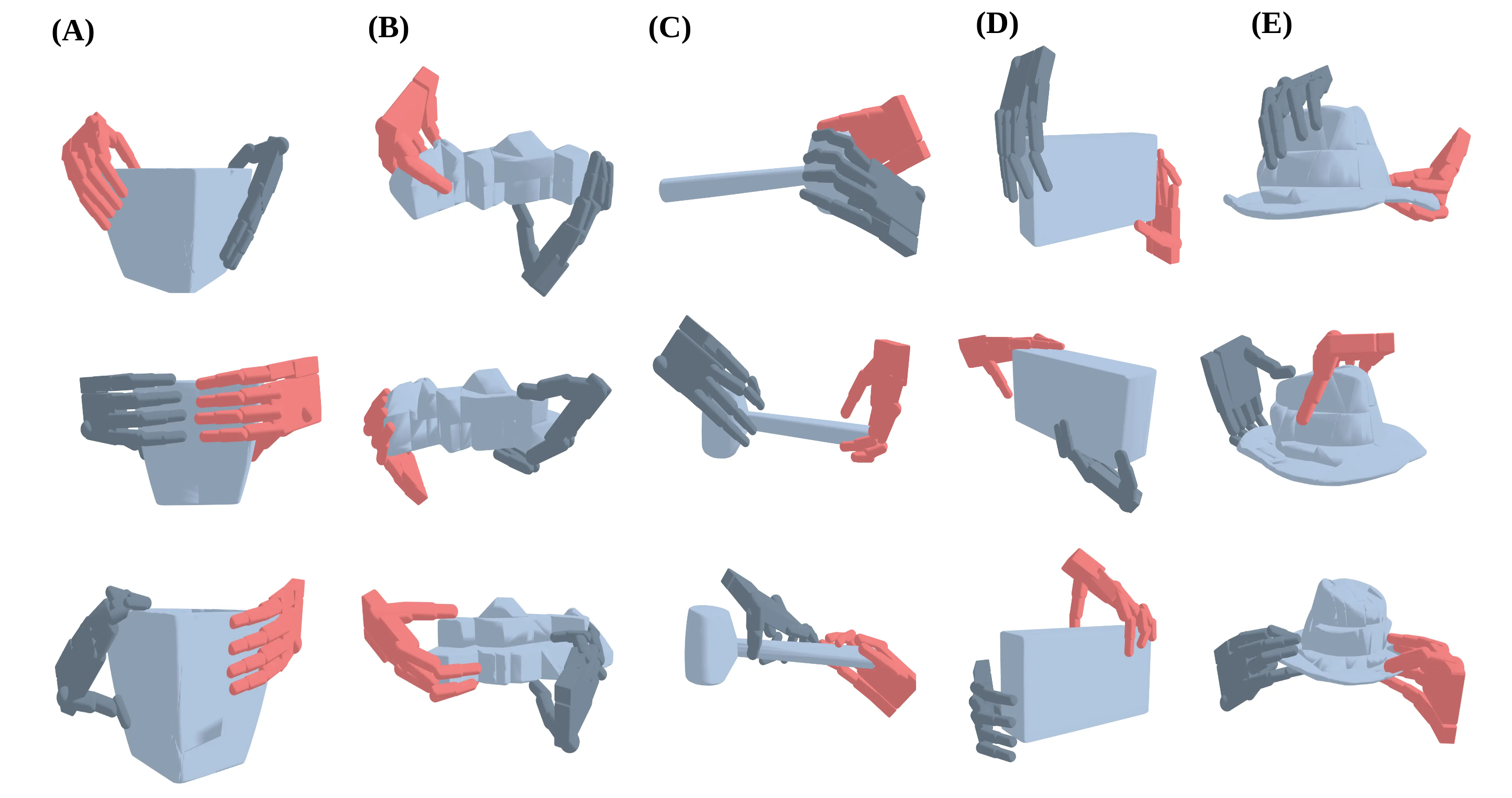}
        \caption{Unseen objects from GSO dataset.}
    \end{subfigure}
    \hfill
    \begin{subfigure}{0.49\textwidth}
        \label{fig:grasps_sub2}
        \centering
        \includegraphics[width=0.97\linewidth]{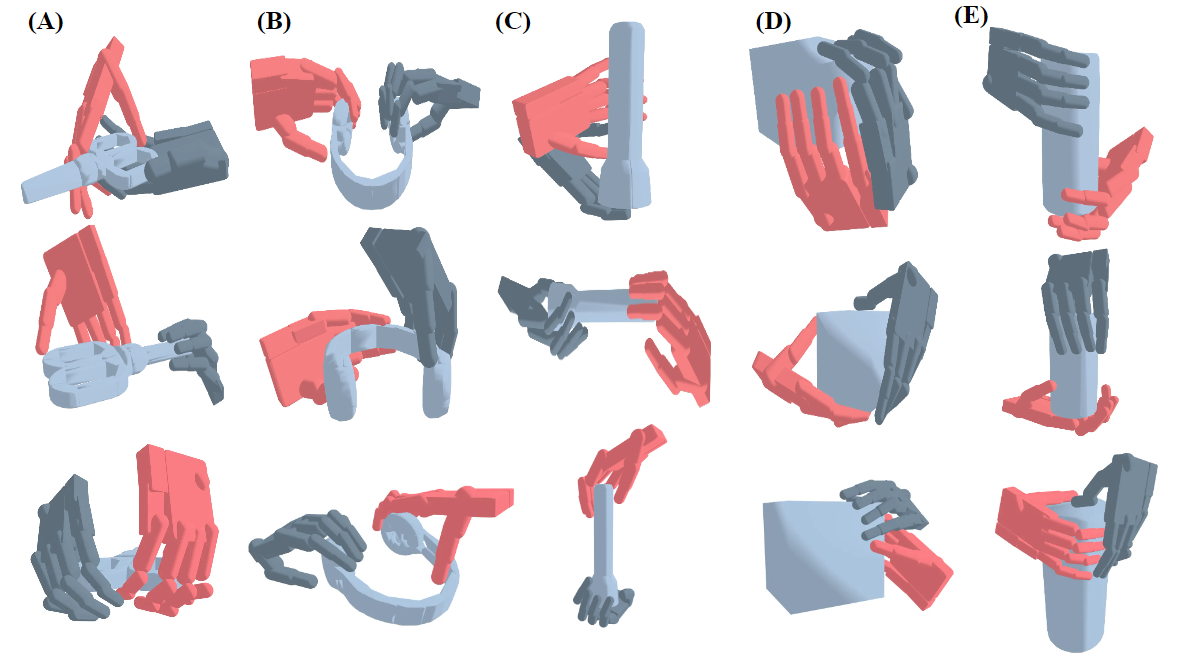}
        \caption{{\revcolor{red}Unseen objects from DDG, YCB, and contactDB datasets.}}
    \end{subfigure}
    \caption{{\revcolor{red} The diverse bimanual stable grasps synthesized with BimanGrasp-DDPM. All objects are unseen during training. } }
     \label{fig:grasps_sub}
\end{figure}

{\revcolor{red} Last, we showcase the generalizability of our model to objects from other datasets. For this, we selected 60 objects from the DDG, YCB, and ContactDB datasets (which differ from the GSO Dataset used for training). We randomly scaled object diameters within the range $[0.2 \text{ m}, 0.4 \text{ m}]$. The object density was fixed at $1000 \text{ kg} \cdot \text{m}^{-3}$. We achieved an average success rate of $63.23\%$. Together with grasp poses generated for GSO objects, the visualizations are shown in Fig.~\ref{fig:grasps_sub}. }

\subsection{Discussions} \label{exp:discussions}

\textbf{Limitations}. While we have demonstrated the effectiveness of our BimanGrasp-DDPM model, the DDPM's grasp synthesis process does not explicitly consider penetration. Consequently, the generative model can still produce infeasible grasp poses. This issue is currently mitigated through a post-processing step. As previously described in Sec.~\ref{method:sec_d}. We anticipate that this limitation can be addressed by leveraging more recent diffusion models that account for physical constraints, such as the approaches described in \cite{yuan2023physdiff}. 

{\revcolor{red} In addition, the algorithms may generate grasp poses that are not human-like. Given that our dataset provides a large number of diverse grasps, we believe a possible solution is to automatically select a subset of the dataset using a visual language model scorer and then retrain the DDPM model under a human-like grasp data distribution.}

{\revcolor{red} 
\textbf{Diversity}. The diversity of grasps was evaluated using an entropy metric $H_{mean}$ adapted from \cite{wang2023dexgraspnet,lu2023ugg}. The mean entropy $H_{mean}$ of grasps generated by BimanGrasp algorithm and BimanGrasp-DDPM is $4.39$ and $3.72$, with standard deviations $H_{std}$ of $0.47$ and $0.49$, respectively. Together with the visualization results in Sec.~\ref{exp:ddpm}, this proves that DDPM can generate multi-modal and diverse bimanual grasps.
}

\begin{figure}[t]
    \centering
    \includegraphics[width=\linewidth]{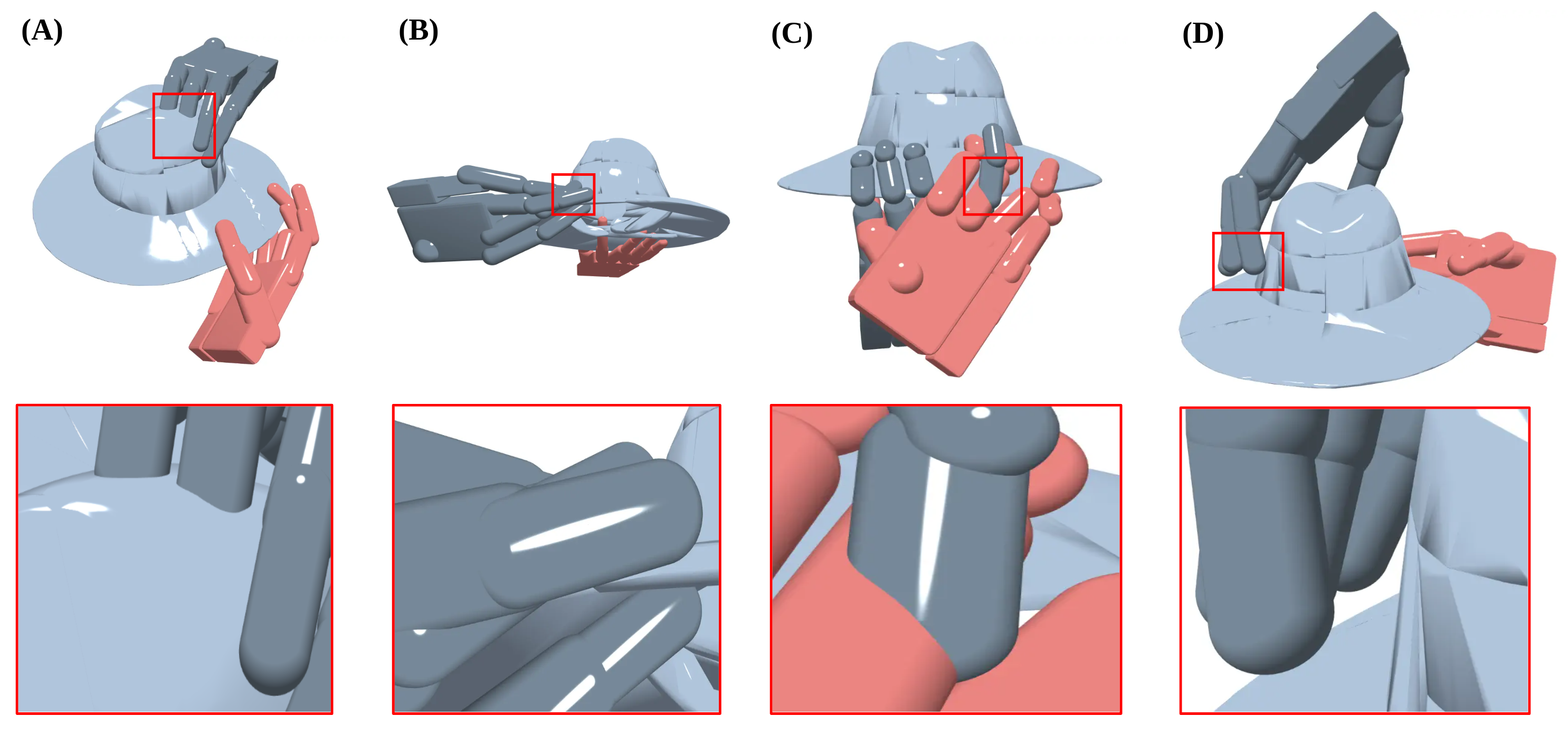}
    \caption{{Visualization of four most common failure patterns: {\revcolor{red}(A) hand-object penetration, (B) hand's self-penetration}, (C) inter-hand penetration, and (D) failure to establish contact.}}
    \label{fig:error_ver2}
\end{figure}

\textbf{Failure Cases}. As part of an ablation study, we highlight failure cases from the BimanGrasp algorithm synthesized grasps in Fig.~\ref{fig:error_ver2}. All these failure patterns could happen both in BimanGrasp algorithm and BimanGrasp-DDPM. Our observations indicate that penetration remains the primary cause of grasp failure. The optimization procedure occasionally fails to retract the hand from objects due to local optima. Additionally, in some instances, the hands may become detached from the objects' surface, causing the object to slip away before finger attached to the object surface.

{\revcolor{red} 
\textbf{Computational Cost}. The synthesis of the BimanGrasp Dataset was achieved on a server with four Nvidia A40 GPUs. It requires 170 GB of GPU memory and takes 117 minutes to generate 4,500 grasps per batch. For DDPM, the inference stage can be accomplished only on a commercial GPU. Using an RTX 4090 GPU, we can parallelize the inference of 64 grasps, with the inference time being 8.19 seconds.

}

\section{Conclusion}

Humans naturally utilize bimanual actions for various object manipulation tasks. However, algorithms for synthesizing bimanual dexterous hands grasping poses were barely studied so far. To bridge this gap, we proposed the BimanGrasp algorithm, which successfully synthesizes bimanual grasps for diverse objects by leveraging stochastic optimization algorithms guided by energy-based heuristics. These grasps are then refined through physical validation in the Isaac Gym environment. Through this process, we obtained the BimanGrasp-Dataset, which contains over 150k verified pairs of grasps for $900$ objects. This dataset enables the training of data-driven models capable of accelerating the grasp synthesis process. To prove this, we developed a BimanGrasp-DDPM model that excels in offering efficient grasp poses and with grasp success rate comparable to that of the BimanGrasp algorithm.

In the future, we plan to further improve the grasp success rate. {\revcolor{red}We also plan to conduct experimental validations of the proposed algorithms using a real-world bimanual humanoid robot}. We believe that the proposed technique can provide impacts to humanoid robots by enhancing their ability to handle various daily life objects in homes and other unstructured environments.

\bibliographystyle{IEEEtran}
\bibliography{references}

\vfill

\end{document}